\documentclass[conference]{IEEEtran}
\IEEEoverridecommandlockouts

\usepackage[font=footnotesize]{caption}
\usepackage{comment}
\usepackage{multicol}
\usepackage{soul}
\usepackage[normalem]{ulem}
\usepackage{multirow}
\usepackage{cite}
\usepackage{stfloats}
\usepackage{amsmath,amssymb,amsfonts}
\usepackage{algorithmic}
\usepackage{lmodern}
\usepackage{xcolor}
\usepackage{graphicx}
\usepackage{textcomp}
\usepackage{subcaption}
\usepackage{rotating}

\usepackage{times}
\usepackage{pgfplots}
\usepackage{hyperref}
\usepackage{orcidlink} 
\usepackage{booktabs}
\pgfplotsset{width=0.99\linewidth,compat=1.9}

\title{Enhancing Apple’s Defect Classification: Insights from Visible Spectrum and Narrow Spectral Band Imaging}

\author{
    \IEEEauthorblockN{\hspace{2cm}}
    \and
    \IEEEauthorblockN{Omar Coello}
    \IEEEauthorblockA{\textit{CIDIS} \\
    \textit{ESPOL Polytechnic University}\\
    Guayaquil, Ecuador \\
    omcoello@espol.edu.ec}
    \and
    \IEEEauthorblockN{Moisés Coronel}
    \IEEEauthorblockA{\textit{CIDIS} \\
    \textit{ESPOL Polytechnic University}\\
    Guayaquil, Ecuador \\
    moagcoro@espol.edu.ec}
    \and
    \IEEEauthorblockN{Darío Carpio}
    \IEEEauthorblockA{\textit{CIDIS} \\
    \textit{ESPOL Polytechnic University}\\
    Guayaquil, Ecuador \\
    dncarpio@espol.edu.ec}
    \and
    \IEEEauthorblockN{\hspace{2cm}}
    \and
    \IEEEauthorblockN{\hspace{2cm}}
    \and
    \IEEEauthorblockN{Boris Vintimilla}
    \IEEEauthorblockA{\textit{CIDIS} \\
    \textit{ESPOL Polytechnic University}\\
    Guayaquil, Ecuador \\
    boris.vintimilla@espol.edu.ec}
    \and
    \IEEEauthorblockN{Luis Chuquimarca}
    \IEEEauthorblockA{\textit{CIDIS} \\
    \textit{ESPOL Polytechnic University}\\
    Guayaquil, Ecuador \\
    lchuquim@espol.edu.ec,\\
    and\\
    \textit{FACSISTEL} \\
    \textit{UPSE Santa Elena Peninsula State University}\\
    La Libertad, Ecuador}
}

\begin{document}
\maketitle
\newcommand{\copyrightnotice}[1]{{%
  \renewcommand{\thefootnote}{}
  \footnotetext[0]{#1}%
}}
\copyrightnotice{979-8-3503-7565-7/24/\$31.00 ©2024 IEEE \newline © 2024 IEEE. Personal use of this material is permitted. Permission from IEEE must be obtained for all other uses, in any current or future media, including reprinting/republishing this material for advertising or promotional purposes, creating new collective works, for resale or redistribution to servers or lists, or reuse of any copyrighted component of this work in other works.

DOI: https://doi.org/10.1109/10677803}


\begin{abstract}
This study addresses the classification of defects in apples as a crucial measure to mitigate economic losses and optimize the food supply chain. An innovative approach is employed that integrates images from the visible spectrum and 660 nm spectral wavelength to enhance accuracy and efficiency in defect classification.
The methodology is based on the use of Single-Input and Multi-Inputs convolutional neural networks (CNNs) to validate the proposed strategies. Steps include image acquisition and preprocessing, classification model training, and performance evaluation.
Results demonstrate that defect classification using the 660 nm spectral wavelength reveals details not visible in the entire visible spectrum. It is seen that the use of the appropriate spectral range in the classification process is slightly superior to the entire visible spectrum. The MobileNetV1 model achieves an accuracy of 98.80\% on the validation dataset versus the 98.26\% achieved using the entire visible spectrum.
Conclusions highlight the potential to enhance the method by capturing images with specific spectral ranges using filters, enabling more effective network training for classification task. These improvements could further enhance the system's capability to identify and classify defects in apples.
\end{abstract}


\section{Introduction}

In today's global market, ensuring the quality control of agricultural products is paramount for both companies and consumers~\cite{qian2020food,han2021comprehensive}. International quality standards are indispensable for the production of high-quality fruits, vegetables, and other agricultural commodities, which are essential for maintaining human health~\cite{mostafidi2020review,saeed2021advances}. The production of high-quality foods such as fruits entails meeting acceptable characteristics (ripeness, deformations, and defects) for both inspectors and consumers~\cite{naranjo2020review,tripathi2020role,chuquimarca2023banana}. The cost of food products is intricately linked to their quality, with categorization by global commercial oversight organizations indicating that higher quality correlates with higher costs. Therefore, for producers, having a system that accurately inspects the quality of their products and identifies their appropriate category is crucial for accessing suitable markets and achieving optimal financial gains.

The identification of fruit defect characteristics is essential to ensure their quality, maintain their nutritional value, satisfy the end consumer, and prevent financial losses for agricultural producers~\cite{pacheco2023fruit}. Currently, manual processes are employed in the agricultural industry to inspect fruit quality~\cite{caldwell2023automation}. These techniques are slow, imprecise, and leave room for the occurrence of defects that lead to fruit rejection by quality inspectors or consumers. To address this issue, machine learning vision techniques have been studied to assess fruit quality externally but more rapidly and accurately.

The present study aims to conduct an evaluation using deep learning architectures such as CNN models for defect detection in apples. The evaluation is carried out using a dataset comprising visible spectrum images and a spectral wavelength of 660 nm as input data to train and validate the models. The defects to be identified include: rot, bruises, scabs, and black spots, which are caused by insects, diseases, climatic conditions, and post-harvest handling. However, according to international quality standards, if a defect is present, the fruit is rejected. Therefore, the classification is binary, meaning that fruits are categorized only as healthy or defective apples.

Our article makes several contributions to the field of apple defect classification:

\begin{itemize}
    \item Acquisition and creation of a dataset comprising three types of apple defects: bruises, stains, and rot, utilizing images captured in both the visible spectrum and the 660 nm wavelength spectrum. Additionally, silhouette images were generated for each defect type. This publicly available dataset on various Apple defects is accessible on GitHub for utilization by the scientific community (\url{https://github.com/cidis-vision/apple-defects}).
    \item Evaluation of the accuracy of Single-Input and Multi-Input neural network architectures for defect type classification using data from different spectral ranges.
    \item Classification of three apple defects for quality control applications.
\end{itemize}

This article is structured as follows: Section \ref{sec:related works} provides a literature review of the works on fruit defect identification. Section \ref{sec:proposed approach} describes the proposed methodology for conducting the study. Section \ref{sec:experimental results} presents the results of identifying healthy and defective apples using publicly available CNN models. Finally, conclusions are presented in Section \ref{sec:conclusions}.

\section{Related Works}
\label{sec:related works}

The surfaces of materials exhibit a unique electromagnetic reflectance, which can be analyzed through the use of hyperspectral imaging~\cite{femenias2021hyperspectral}. In recent years, research has focused on defect detection in fruits using this specific type of imaging, particularly in the infrared wavelength region~\cite{soltani2022defect}. This is attributed to the distinct behavior of reflectance on healthy surfaces compared to those with defects in fruits. Furthermore, several studies have delved into the utilization of hyperspectral imaging across various regions of the infrared spectrum, including near-infrared (400-1000 nm), short-wave infrared (1000-2500 nm), and mid-wave infrared (3500-5000 nm)~\cite{li2018detection,an2023advances}. In our case, we conducted a Narrow-band spectral analysis, as we utilized a narrow-band filter centered at a specific wavelength (660 nm) with a bandwidth of 60 nm. This refers to a window in the electromagnetic spectrum that is being examined selectively and specifically.

Several studies have demonstrated that images of fruits captured in the Narrow-band spectrum offer superior results in defect identification compared to other ranges within the infrared spectrum~\cite{li2018detection,zhu2016predicting,arango2021quality}.

In~\cite{arango2021quality}, the architectures VGG19, Resnet50, and DesNet121 were evaluated for bruise detection in apples. The best result obtained in this study with RGB images was achieved using the DenseNet121 model with an average Matthews Correlation Coefficient (MCC) of 0.978. Conversely, the Resnet50 model achieved the highest average MCC value with near-infrared reflectance (NIR) images at 0.97. After analyzing the results, a model fusion technique with multispectral images using RGB and NIR images could be employed, combining the outcomes across different spectral bands to obtain more robust CNN models. However, the identification of other defects such as stains, rot, and black spots is not considered to enhance the generalizability of the CNN model.

A detailed overview of the utilization of computer vision technology combined with uniform and structured illumination for fruit defect detection is provided in~\cite{lu2018detection}, specifically focusing on 'Delicious' and 'Golden Delicious' apples. It is demonstrated that the multispectral imaging technique is effective in detecting both surface and subsurface defects in apples, and various machine learning algorithms are compared to enhance detection accuracy. Furthermore, the employment of a deep learning approach, specifically a CNN model, exhibits the most notable improvement in defect detection accuracy, achieving 98\% for both varieties of apples. Although a wide variety of defects in apples is considered (such as Bitter Pit, Frost Ring, Russeting or Scab, Lenticel Spots or Breakdown, Mechanical Injury, Insect Damage, Rot or Decay, Fecal Contamination), the system only classifies Bitter Pit, Bruise, subsurface injuries, Scab, Frost Ring, and Rot. The number of images per defect is minimal to ensure that a CNN model is optimal and generalizable.

The study~\cite{feng2019detection} presents a comprehensive analysis of the use of hyperspectral imaging for the detection of subtle bruises in winter jujubes. Through the application of NIR and visible/near-infrared reflectance (Vis-NIR) hyperspectral imaging techniques, researchers conducted pixel-wise spectral analysis and employed machine learning algorithms for classification. Notably, CNN models demonstrated superior performance in bruise detection, especially when utilizing Vis-NIR spectra. The comparison of models based on SVM, LR, and CNN revealed CNN's effectiveness with 100\% accuracy in feature extraction, particularly in scenarios with multiple interference factors. Therefore, the possibility arises of applying CNN models with images in other spectra besides the visible, such as NIR, to detect various defects in different fruits. In our study, we focused on defects present in apples.

\section{Proposed Approach}
\label{sec:proposed approach}
In this section, the dataset acquisition and the proposed strategy are presented. Our proposed approach aims to improve defect classification accuracy in apples, specifically focusing on the identification of bruises, stains, rot, and black spots. Leveraging the power of CNNs, we propose two strategies to refine the accuracy of defect classification. The significance of this endeavor lies in its potential to bolster the efficiency of postharvest quality control in the fruit industry.


\subsection{Dataset Acquisition}
The dataset used in this study was obtained through a specialized acquisition system employing two parallel cameras for simultaneous captures. The first camera employed was the MER2-160-227U3C, designed for capturing images in the visible spectrum. The second camera utilized was the Basler ace acA1300-60gmNIR, equipped with a bandpass filter BP660 with a useful range between 640-680nm, centered at 660 nm, and a Full Width at Half Maximum (FWHM) of 60 nm. The camera's resolutions are 1280x960 pixels and 1280x1024 pixels respectively.

To ensure comprehensive coverage of the entire fruit, a rotating table was employed where the fruit was positioned. This setup facilitated the capture of 120 images per fruit. The synchronized operation of both cameras allowed for simultaneous acquisition, ensuring temporal coherence in the dataset.

The captured dataset encompasses three types of defects: bruises, stains, and rot. Given the simultaneous nature of the captures, an equal number of images were obtained for both cameras. To specifically address the defect of bruises, six apples in good condition were subjected to mild impacts, and captures were conducted over six days. For the defect of stains, 20 apples were utilized, and for rot, another set of 20 apples was used. This targeted approach ensured the inclusion of realistic variations in the dataset, allowing the models to learn and generalize well to different instances of these defects. 

The total number of captured images for each defect category was as follows: 4539 for bruises, 3136 for stains, and 3094 for rot, resulting in a comprehensive dataset comprising a total of 10769 images. This diverse dataset, with varied angles and defect types, is fundamental for training and evaluating the proposed defect classification models, ensuring their effectiveness across different scenarios and conditions. Before training the models, 2 stages of image post-processing were carried out; image registration and defect segmentation.

\subsubsection{Image Registration}
Following the acquisition of images, the image registration stage became imperative due to the configuration of the dual-camera acquisition system. A slight spatial misalignment existed between the two cameras, necessitating the registration of images to align the image pairs accurately. In this step, the LightGlue \cite{lindenberger2023lightglue} registration model was employed, yielding satisfactory results in achieving spatial coherence.

The registration process aimed to rectify any misalignment introduced during simultaneous captures, ensuring accurate correspondence between corresponding pixels in the paired images.

\begin{figure}[!t]
  \centering
  \includegraphics[scale=0.6]{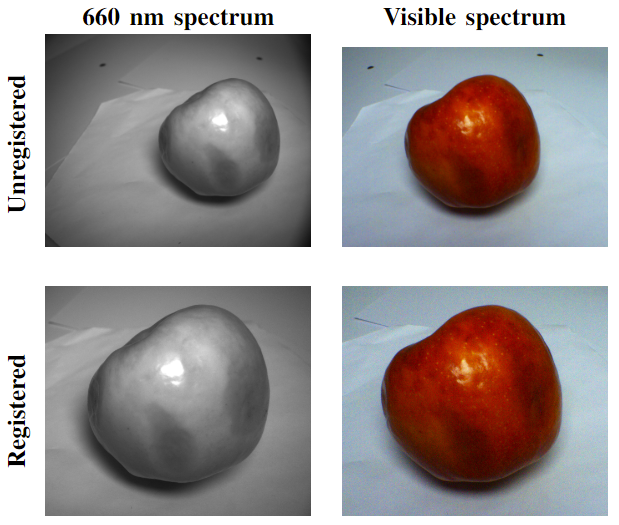}
  \caption{Image registration stage. The images in the first row show unregistered images while those in the second row show the resulting registered images.}
  \label{fig:image_registration}
\end{figure}

Moreover, as part of the image registration pipeline, the images underwent a cropping process to standardize their resolution to 960x830 pixels, Figure \ref{fig:image_registration} shows the resulting images from the registration process. This standardization facilitated consistency in subsequent processing steps and ensured that all images had a uniform resolution for effective feature extraction and defect detection.

The combination of image registration and resolution standardization enhances the overall coherence and comparability of the dataset, laying the foundation for subsequent stages in the defect detection process. The alignment achieved through image registration mitigates spatial variations, allowing for a more accurate analysis of defects and improving the robustness of the subsequent machine-learning models.

\subsubsection{Defects Segmentation} 
After obtaining the registered datasets at a standardized resolution, the subsequent step involved the segmentation of defects. The objective was to automatically generate segmentation masks that describe the regions affected by the defects. To streamline this process, the Segmentation Anything Model \cite{kirillov2023sam} was employed for inference, utilizing pre-trained weights provided by the authors. In this context, no model training occurred, and parameters were adjusted specifically for inference to accept very small regions, such as stains.

The model was used solely for inference, the parameters were carefully adjusted to ensure the model recognized and delineated even small regions that might represent defects like stains. The focus was on optimizing the model's sensitivity to capture features associated with bruises, stains, and rot.

Following the inference step, post-processing techniques were applied to refine the segmentation results. This involved filtering and preserving only the relevant regions that accurately described the targeted defects. The post-processing step aimed to maintain high precision in defect localization, ensuring that the segmentation masks provided valuable and accurate spatial information for subsequent defect classification models.

\subsection{Spectral Subrange Selection}
We hypothesized that focusing on a specific subrange of the visible spectrum could enhance defect visibility. To test this, we selected a wavelength of 660 nm with a bandwidth of 60 nm. 

The wavelength of 660 nm falls within the red part of the visible spectrum. In this range, apples generally exhibit strong absorption characteristics, particularly due to pigments such as anthocyanins and chlorophyll. The red color is particularly sensitive to changes in the physiological condition of the fruit, making it an ideal choice for defect classification. Anthocyanins, responsible for the red coloration in apples, can undergo alterations in the presence of physiological stress, diseases, or decay, leading to observable changes in color.

By focusing on the 660 nm wavelength, we leverage the sensitivity of this spectral range to subtle variations in apple coloration. Rot, stains, bruises, and other defects often manifest as alterations in pigmentation, affecting the overall appearance of the fruit. The use of 660 nm allows us to capture these color changes with precision, enabling the classification of defects at an early stage when visual cues might not be apparent to the human eye.

Furthermore, the choice of 660 nm aligns with the practical considerations of imaging technology. Cameras and sensors optimized for this wavelength can provide high-resolution images, facilitating the extraction of detailed information for defect classification. This targeted approach enhances the efficiency and accuracy of the defect classification process, contributing to the overall effectiveness of postharvest quality control. We trained and evaluated our models using both datasets:

\begin{itemize}
    \item Visible Spectrum Dataset: Images captured across the entire visible spectrum.
    \item 660 nm Spectrum Dataset: Images captured specifically at the selected wavelength.
\end{itemize} 

By comparing the model performance on these datasets, we aimed to validate whether the spectral subrange improves defect classification.

\subsection{Classification Models}
\begin{figure*}[!t]
  \centering
  \includegraphics[scale=0.45]{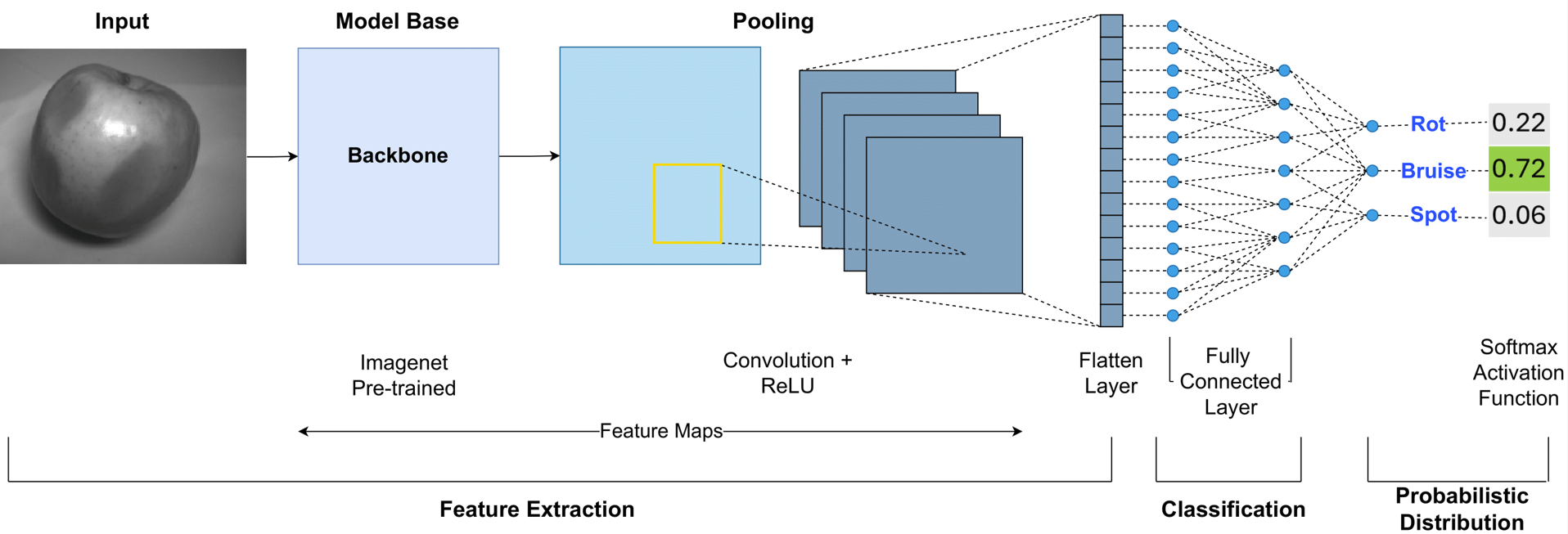}
  \caption{Single-Input Architecture for apple defects classification.}
  \label{fig:feedforward}
\end{figure*}

The study employs well-established object classification models from the state-of-the-art as a backbone, including DenseNet121, VGG19, MobileNetV1, and ResNet50. These models are widely recognized for their effectiveness in computer vision tasks. To tailor them to the task of classifying apples into three distinct categories bruises, stains, and rot, we implement a series of architectural refinements. These refinements include a strategic integration of a pooling layer, followed by two densely connected layers, each featuring a dropout mechanism set at 0.5 to mitigate overfitting. The final layer comprises a dense output layer housing three nodes that encode the probabilities associated with each class. Adam optimizer was used on the training stage with a learning rate of 0,0001. This refined architecture, depicted in Figure \ref{fig:feedforward}, was employed for the four backbone models, these backbone models were used with the pre-trained weights on Imagenet as feature extractors. 

\subsection{Multi-Inputs Architecture with Segmentation Masks}
\begin{figure*}[!t]
  \centering
  \includegraphics[scale=0.45]{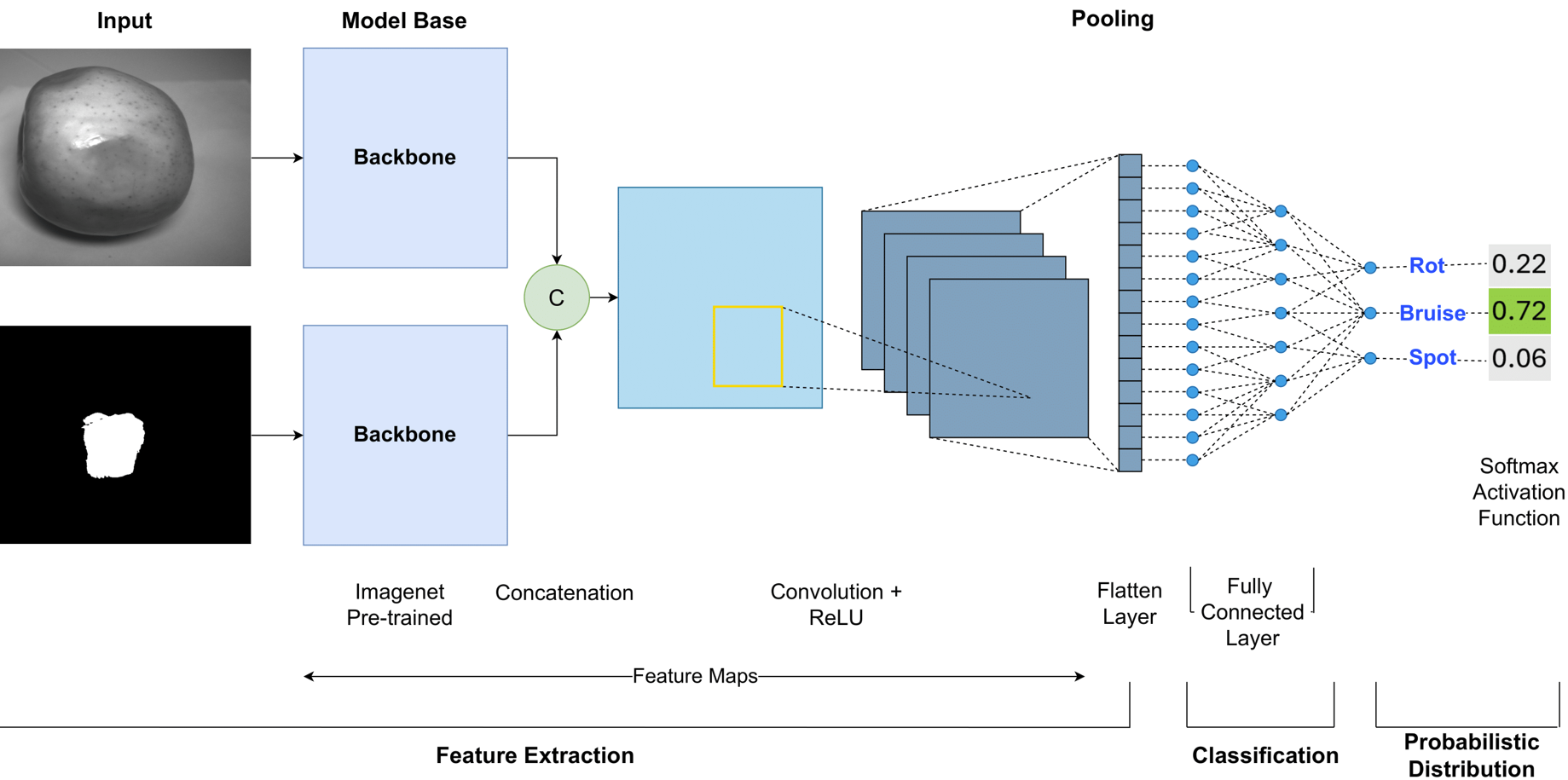}
  \caption{Multi-Inputs Architecture for apple defects classification.}
  \label{fig:siamese}
\end{figure*}

Our second hypothesis was that incorporating segmentation masks of defects as additional input could enhance classification accuracy. To create the segmentation masks, we employed the SegmentAnything model, which generated pixel-wise masks for each defect type. Using the same backbone models mentioned earlier (DenseNet121, VGG19, MobileNetV1, and ResNet50), we implemented a Multi-Inputs architecture. This architecture takes both the original apple image and its corresponding defect segmentation mask as input as shown in Figure \ref{fig:siamese}. Both images are processed in each branch by the backbone model, then the resulting feature vectors are concatenated along the depth dimension to finally be processed by the classifier. The Multi-Inputs network learns to associate the features of the image and its mask, improving the model’s ability to discriminate defects.

\section{Experimental Results}
\label{sec:experimental results}

\subsection{Single-Input Method}
Using pre-trained weights from ImageNet allowed the models to achieve swift convergence at around 25 epochs. Each model was exposed to two training sessions: one applying the 660 nm spectrum dataset and the other using the visible spectrum dataset. Table \ref{tab:accuracy} presents the accuracy percentages achieved with the classification models on the validation sets. The highest-performing model was MobileNetV1, achieving over 98\% precision in both datasets, highlighting MobileNetV1's ability to efficiently handle Single-Input, low-resolution images was a significant advantage when processing multispectral images, potentially leading to improved accuracy in defect classification in apples. That model performed better in the 660 nm spectrum dataset than the visible spectrum dataset, recording the respective precision rates of 98.8\% and 98.26\%. 

\subsection{Multi-Inputs Method}
In our investigation, three experiments were performed to assess different configurations and input modalities.

For the first experiment, we utilized the 660 nm spectrum dataset along with its respective defect segmentation masks as inputs. Among the tested models, MobileNetV1 achieved the highest precision, reaching 75.14\%.

In the second experiment, we employed the visible spectrum dataset along with its corresponding defect segmentation masks. Notably, DenseNet121 emerged as the top-performing model, attaining a precision rate of 85.71\%, as illustrated in Table \ref{tab:accuracysiamese1}. By establishing dense connections between layers, DenseNet121 promotes feature reuse and information flow throughout the network, enabling more efficient gradient propagation and feature extraction. This characteristic fosters deep representations of complex patterns inherent in apple defect images, ultimately enhancing classification accuracy. Moreover, DenseNet121's dense connectivity mitigates the vanishing gradient problem, facilitating effective training of deeper networks and enabling the model to learn intricate relationships among features, which is particularly beneficial in discerning subtle differences between various types of apple defects.

The third experiment involved using the 660 nm spectrum dataset in conjunction with the corresponding images from the visible spectrum dataset. Here, MobileNetV1 exhibited superior performance with a precision of 90.91\%. Detailed results for the third experiment are provided in Table \ref{tab:accuracysiamese2}.

These models better leveraged transfer learning of weights pre-trained on the ImageNet dataset, improving their ability to discern relevant features from apple defect images. Additionally, regularization techniques such as normalization and batch dropout helped avoid overfitting, ensuring robust performance on unseen data.

Compared to the Single-Input approach results detailed in Table \ref{tab:accuracy}, the accuracy of all models exhibited a decline across both spectra, except for the VGG19 model. Remarkably, employing a dual-input approach with defect segmentation masks doubled the accuracy specifically for VGG19 from about 33\% to 65-66\%. This observation hints at a complex interplay among model architecture, feature extraction methods, and dataset characteristics. The default concatenation strategy, particularly along the depth dimension, seemingly favored feature integration within the VGG19 model over others. This concatenation method assumes that features from both branches are mutually compatible and complementary, a condition that may not hold universally across all models. The exceptional performance of VGG19 in the dual-branch setup could be attributed to its inherent robustness in feature representation. In contrast, models like MobileNetV1, DenseNet121, and ResNet50, while proficient in feature extraction, may have struggled to adapt their architectures and feature representations to effectively leverage the dual-branch approach within this specific experimental context.

Upon examining the results presented in Table \ref{tab:accuracysiamese2}, it becomes apparent that combining both spectra does not yield a precision enhancement over training each dataset individually. However, there's a marginal 3\% improvement with the VGG19 model, this reaffirms the adaptability of the VGG19 model to capitalize on the integration of multiple inputs for enhanced feature representation.

\begin{table}
\centering
\caption{Accuracy values of classification models on Single-Input Architecture for 660 nm and visible spectrum datasets.}
\begin{tabular}{|l|c|c|}
\hline
\textbf{Model} & \textbf{660 nm spectrum (\%)} & \textbf{Visible spectrum (\%)} \\ \hline
MobileNetV1        & 98.8                               & 98.26 \\ \hline
DenseNet121        & 95.39                               & 98.46 \\ \hline
ResNet50        & 95.79                               & 96.10 \\ \hline
VGG19       & 33.36                               & 33.22 \\ \hline
\end{tabular}

\label{tab:accuracy}
\end{table}

\begin{table}
\centering
\caption{Accuracy values of classification models on Multi-Inputs Architecture for 660 nm spectrum + defect masks and visible spectrum + defect masks datasets.}
\begin{tabular}{|l|c|c|}
\hline
\multirow{2}{*}{\textbf{Model}} & \textbf{660 nm spectrum +} & \textbf{Visible spectrum +}  \\
& \textbf{defect masks (\%)} & \textbf{defect masks (\%)} \\ \hline
MobileNetV1        & 75.14  & 69.85 \\ \hline
DenseNet121        & 72.59 & 85.71 \\ \hline
ResNet50        & 35.63 & 52.87 \\ \hline
VGG19       & 65.78  & 66.04 \\ \hline
\end{tabular}

\label{tab:accuracysiamese1}
\end{table} 

\begin{table}
\centering
\caption{Accuracy values of classification models on Multi-Inputs Architecture using 660 nm + visible spectrum datasets.}
\begin{tabular}{|l|c|}
\hline
\textbf{Model} & \textbf{660 nm + visible spectrum (\%)} \\ \hline
MobileNetV1        & 90.91\\ \hline
DenseNet121        & 76.70\\ \hline
ResNet50        & 60.36\\ \hline
VGG19       & 36.36\\ \hline
\end{tabular}

\label{tab:accuracysiamese2}
\end{table}

\section{Conclusions}
\label{sec:conclusions}
In the analysis employing the Single-Input architecture, the classification model achieving the highest precision demonstrated a notable improvement of 0.54\% when trained with the 660 nm spectrum dataset, achieving a precision rate of 98.8\%. This reaffirms the effectiveness of this spectral range for the task of classifying apple defects encompassing bruising, staining, and rot.
However, when employing the Multi-Inputs architecture and utilizing defect segmentation masks provided by the SAM model, no discernible improvement in precision was observed. This could be attributed to the inherent inaccuracies within the defect segments, leading to false positives and false negatives. This issue may be mitigated by refining the segmentation masks obtained with SAM before integration into the training pipeline.
Lastly, employing images from both the 660 nm and visible spectrum datasets as inputs yielded better performance than using defect segmentation masks. Nonetheless, this combined approach failed to surpass the individual performance achieved by the Single-Input scheme. Thus, we conclude that combining both spectra in the mMlti-Inputs scheme does not contribute to the precision of apple defect classification. For future research, it is recommended to consider a finer segmentation of the spectrum in the dataset images, including the incorporation of the NIR spectrum. This approach will enable the evaluation of whether the models' performance is sustained or enhanced with the inclusion of these additional spectral bands.

\section*{Acknowledgements}
This work has been partially supported by the ESPOL-CIDIS-11-2022 project.

\bibliography{mybibfile}

\end{document}